\documentclass[10pt,twocolumn,letterpaper]{article}

\usepackage{authblk}
\usepackage[flushmargin]{footmisc}

\usepackage[pagenumbers]{cvpr}
\usepackage{multirow}
\usepackage{array}
\usepackage{bbm}
\usepackage{graphicx}
\usepackage{amsmath}
\usepackage{amssymb}
\usepackage{booktabs}
\usepackage{bbding}
\usepackage{color}
\usepackage{appendix}
\usepackage{fancyhdr}
\usepackage{CJKutf8}

\usepackage[dvipsnames]{xcolor}

\definecolor{cvprblue}{rgb}{0.21,0.49,0.74}
\usepackage[pagebackref,breaklinks,colorlinks,citecolor=cvprblue]{hyperref}
\usepackage{xcolor}
\definecolor{DarkGreen}{rgb}{0.43, 0.68, 0.28}

\newcommand\keywords[1]{\textbf{Keywords}: #1}

\title{MOSS: Motion-based 3D Clothed Human Synthesis from Monocular Video}

\author[1,2]{Hongsheng Wang\textsuperscript{*}}
\author[2]{Xiang Cai\textsuperscript{*}}
\author[2]{Xi Sun}
\author[2]{Jinhong Yue}
\author[2]{Zhanyun Tang}
\author[1]{Shengyu Zhang\textsuperscript{$\dag$}} 
\author[2]{\\ Feng Lin}
\author[1]{Fei Wu}

\affil[1]{Zhejiang University, China}
\affil[2]{Zhejiang Lab, China}

\makeatletter
\renewcommand\AB@affilsepx{, \protect\Affilfont}
\makeatother

\begin{document}
\begin{CJK}{UTF8}{gbsn}
\maketitle

\lfoot{This work has been submitted to the IEEE for possible publication. Copyright may be transferred without notice, after which this version may no longer be accessible.}

\renewcommand{\thefootnote}{\fnsymbol{footnote}}
\footnotetext[1]{These authors contributed equally to this work.}
\footnotetext[2]{Corresponding Author.}
\renewcommand\thefootnote{}
\footnote{This work has been submitted to the IEEE for possible publication. Copyright may be transferred without notice, after which this version may no longer be accessible.}
\begin{abstract}

Single-view clothed human reconstruction holds a central position in virtual reality applications, especially in contexts involving intricate human motions. It presents notable challenges in achieving realistic clothing deformation. Current methodologies often overlook the influence of motion on surface deformation, resulting in surfaces lacking the constraints imposed by global motion. To overcome these limitations, we introduce an innovative framework, \textbf{M}otion-Based 3D Cl\textbf{o}thed Humans \textbf{S}ynthe\textbf{s}is (MOSS), which employs kinematic information to achieve motion-aware Gaussian split on the human surface. Our framework consists of two modules: \textbf{K}inematic \textbf{G}aussian Loc\textbf{a}ting \textbf{S}platting (KGAS) and S\textbf{u}rface Deformat\textbf{i}on \textbf{D}etector (UID). KGAS incorporates matrix-Fisher distribution to propagate global motion across the body surface. The density and rotation factors of this distribution explicitly control the Gaussians, thereby enhancing the realism of the reconstructed surface. Additionally, to address local occlusions in single-view, based on KGAS, UID identifies significant surfaces, and geometric reconstruction is performed to compensate for these deformations. Experimental results demonstrate that MOSS achieves state-of-the-art visual quality in 3D clothed human synthesis from monocular videos. Notably, we improve the Human NeRF and the Gaussian Splatting by 33.94\% and 16.75\% in LPIPS* respectively. Codes are available at \url{https://wanghongsheng01.github.io/MOSS/}.

\end{abstract}
\vspace{-2mm}
\section{Introduction}

\begin{figure*}[t]
    \centering

    \includegraphics[width=0.8\linewidth]{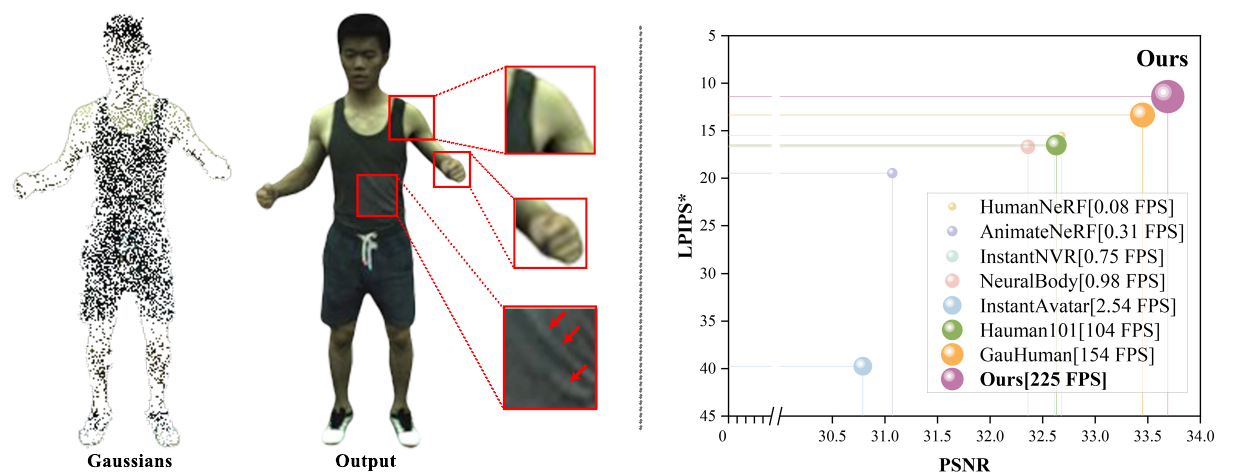}
    \caption{
     MOSS reconstructs 3D clothed humans with detailed joints and fine clothing folds. The right image demonstrates that MOSS surpasses the visual quality of previous works on MonoCap. (LPIPS* = LPIPS × 10$^{3}$). Larger circles denote higher FPS.
    }
    \label{fig:fig1}
 
\end{figure*}

\label{sec:intro} 
The significance of accurately modeling clothed humans is on the rise, especially in industries such as fashion, entertainment, and computer games~\cite{Su2020MulayCapMH,Huang2020ARCHAR,Bhatnagar2020LoopRegSL}. When striving to create lifelike mannequins, one of the most significant challenges lies in accurately depicting the dynamic movements of the human body.

Recently, 3D Gaussian Splatting (3DGS)~\cite{Franke2024TRIPSTP,Bolanos2024GaussianSC,Lan20232DGuided3G,Qian20233DGSAvatarAA,Kocabas2023HUGSHG,Cen2023SegmentA3,Lin2023GaussianFlow4R} has greatly improved the efficiency and quality of 3D reconstruction by sampling object geometry through Gaussian distributions. Existing work on reconstructing the human body with 3DGS~\cite{Hu2023GauHumanAG, Li2023Human101T1} introduces the SMPL~\cite{Loper_Mahmood_Romero_Pons-Moll_Black_2015} as the human body prior, which can restore a more realistic human body. However, these works neglect the hierarchical structure constraints of the kinematic tree~\cite{Wu_2020} and the constraints of global motion information in reconstructed human surface, resulting in details missing in the joints of the moving human body and the surface parts of tight-fitting clothing. Another line of works~\cite{Xiu2021ICONIC, Xiu2022ECONEC} developed normals to model clothing to characterize the deformation superficially. The influence of mutual occlusion between clothing folds on the fold rendering was not considered, resulting in the loss of clothing folds on the reconstructed human surface.

To confront the present difficulties, we introduce an innovative framework \textbf{M}otion-Based Cl\textbf{o}thed 3D Humans \textbf{S}ynthe\textbf{s}is (MOSS). MOSS commences by examining the origin of the surface deformation. We use the motion factors (displacement and rotation) from the kinematic tree for Gaussian control to improve the human body reconstruction in large-scale motion.

To effectively tackle the issue of detail blurring from human body movement, we propose KGAS. This module links global motion with Gaussian point shapes, using the matrix-Fisher~\cite{DS,Khatri1977TheVM,downs1972orientation} probability distribution of human body joints to derive rotation factors and principal axis concentrations. The body pose is parameterized by relative 3D joint rotations along the SMPL kinematic tree, which can be represented using rotation matrices. However, regressing these rotation matrices using neural networks presents a challenge because they belong to SO(3)~\cite{lee2018bayesian,mohlin2020probabilistic}, a non-linear 3D manifold with a different topology than the unconstrained outputs of neural networks. Despite this difficulty, it is possible to define probability density functions over the Lie group SO(3). One such function is the matrix-Fisher distribution, whose parameters can be easily regressed by a neural network~\cite{Mohlin2020ProbabilisticOE,Sengupta2021HierarchicalKP}. We use a hierarchical probability distribution over relative 3D joint rotations along the SMPL kinematic tree, where the probability density function of each joint's relative rotation matrix is a matrix-Fisher distribution conditioned on the parents of that joint in the kinematic tree. We train a deep neural network to predict the parameters of such a distribution over body pose, along with a Gaussian distribution over SMPL shape. By globally associating these factors and controlling Gaussians for 3DGS rendering, we achieve precise surface depiction. The principal axis concentration in the Gaussian layout process adjusts sampling probabilities for capturing surface deformations accurately. Through dynamic adjustments in direction and scale, the rotation factor and principal axis concentration ensure each Gaussian faithfully represents the human body's surface deformation under global motion constraints.

Expanding on KGAS, we introduce UID to tackle clothing wrinkle loss caused by limited 2D image supervision due to local occlusion. By evaluating the degree of directional change in the local distribution of Gaussians through comparing the angles between the normal vectors of adjacent Gaussian points with a set threshold, UID can identify and densely process regions on the surface that exhibit significant deformation. In these regions, we employ the KGAS method for geometric reconstruction compensation. This facilitates the restoration of obscured structural details.

Based on the synergistic work of KGAS and UID, our method can improve the reconstruction quality of the clothed human in scenes with substantial human motion, while ensuring the efficiency of training and rendering. Our contributions are as follows:

(1) MOSS creatively extracts global motion factors from the origin of surface deformation. Employing these factors to guide Gaussians to the finest granularity restores realistic joint details and fine clothing folds in large-scale motion.

(2) We propose KGAS to propagate motion information obtained from the kinematic tree to all Gaussians, which effectively guides the 3D Gaussians, accurately restoring the surface deformation.

(3) We propose UID to detect variations of the orientation of local Gaussians, identifying large deformations to articulate the geometric deformation of the human body surface in a directional manner.

(4) Experimental results show that MOSS achieves state-of-the-art human visual perception performance with real-time rendering.

\section{Related Work}
\begin{figure*}[t]
    \centering
     
    \includegraphics[width=1.0\linewidth]{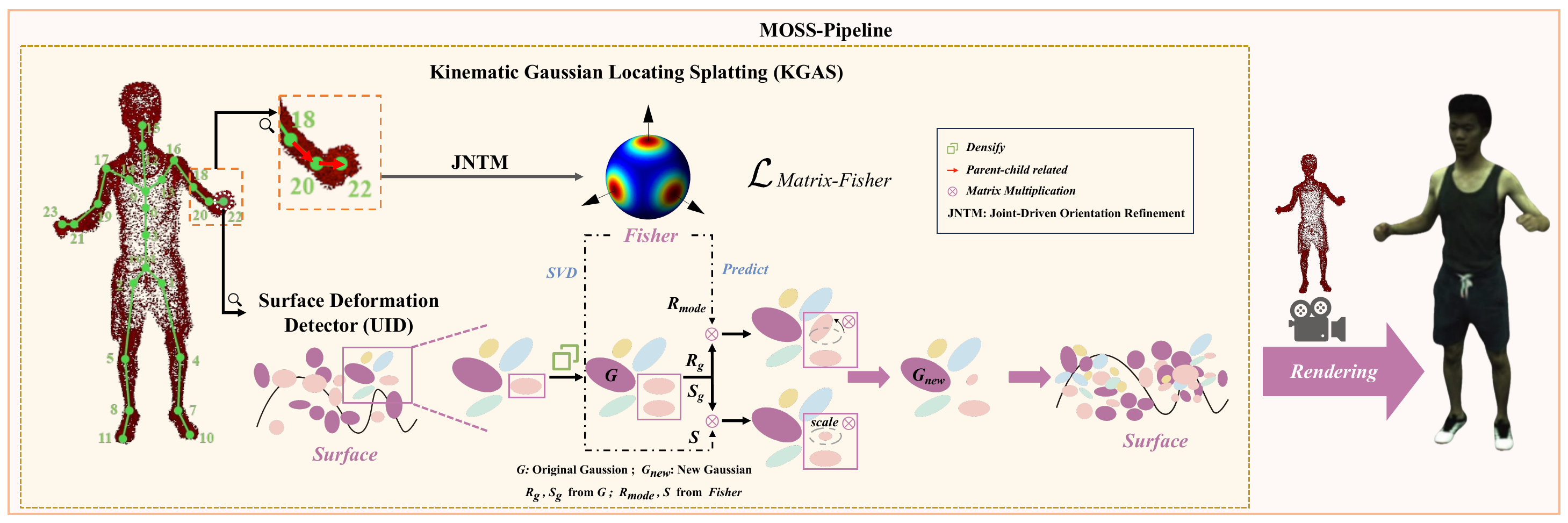}
    \caption{\textbf{MOSS framework.} \textbf{Moss} conditions the Fisher distribution of child joints on the Fisher distribution of their parent joints within the kinematic hierarchy tree, thereby linking the rotational matrices of each joint to the global motion by Joint-Driven Orientation Refinement. The \textbf{UID} is employed to detect and locate areas with numerous surface folds on the human body. In these regions, the Gaussians are scaled by the axial matrix from the SVD of the Fisher and rotated by the directional matrix predicted from the Fisher using \textbf{KGAS}. The T-pose is then converted to the target pose, and the surface folds are refined accordingly.}
    \label{fig:fig2}
\end{figure*}

\subsection{3D Reconstruction and Rendering}
In the domain of 3D reconstruction and rendering, NeRF~\cite{Mildenhall2020NeRFRS,Chen2024Sketch2NeRFMS,Kim2024GARFieldGA,Song2024Tri2planeVA,Nakayama2024ProvNeRFMP,Ming2024HighQualityMB,Zhao2022HumanPM,Yan2024ForgingVF,Metzer2022LatentNeRFFS,JunSeong2022HDRPlenoxelsSH,Gu2023NerfDiffSV} has been recognized for its ability to enable high-quality rendering. However, NeRF often requires significant time and resources for computation. Recently, 3D Gaussian Splatting~\cite{Franke2024TRIPSTP,Bolanos2024GaussianSC,Lan20232DGuided3G,Qian20233DGSAvatarAA,Kocabas2023HUGSHG,Cen2023SegmentA3,Lin2023GaussianFlow4R} has emerged as a promising method for point cloud rendering. By incorporating a spherical harmonic function to represent color attributes and facilitating efficient computation, 3DGS enhances the perception of geometric structures through pixel-level Gaussians. This advancement significantly enhances the efficiency and quality of 3D reconstruction processes.

\subsection{Human Reconstruction}
In the field of human modeling, many methods have been used to model human shape and pose. One commonly used method is SMPL, which is a vertex-based additive human body model that uses linear algebraic models to separately model shape and pose. The SMPL model employs Linear Blend Skinning (LBS)~\cite{lewis2000matrix} technology to smoothly deform the 3D mesh based on the influence of multiple bones in a skeletal animation system, thereby achieving efficient and realistic human pose modeling.

At this stage, based on 3DGS, some methods~\cite{Bogo2016KeepIS,Chen20223DHB,Dittadi2021FullBodyMF,Madadi2020SMPLRDL} use an explicit human structure SMPL priori to aid in the 3DGS optimization process to reconstruct diverse and realistic 3D human models with high-quality. PSAvatar~\cite{Zhao2024PSAvatarAP} uses discrete geometric primitives to create parametric deformable shape models that are represented in fine detail and rendered with high fidelity with the help of 3D Gaussian. Human101~\cite{Li2023Human101T1} uses human-centered forward in the SMPL prior to deforming 3D human models. Gaussian animation method to deform the parameters of 3D Gaussian and improve the rendering speed. GauHuman~\cite{Hu2023GauHumanAG} not only introduces SMPL prior, but also further accelerates the 3DGS by introducing KL dispersion. These methods can restore a more realistic human appearance. However, they lack the constraints of the SMPL kinematic tree when considering human joint linkage effects and surface deformation, which can lead to problems such as blurred joint edges or confusion between joints and clothing when reconstructing complex human poses or when there is clothing occlusion.

\subsection{Surface Reconstruction}
In the realm of reconstructing human clothing surfaces, traditional template mesh models have limitations in capturing clothing details, some approaches use neural representations~\cite{Saito2019PIFuPI,Saito2020PIFuHDMP} to handle complex details like hairstyles and clothing, which considers the human body surface as an implicit function, while utilizing neural networks to learn parameters and effectively model the complex details. Other~\cite{Xiu2021ICONIC,Xiu2022ECONEC} specifically utilize normals to handle extreme surface poses. Recent advances~\cite{Mildenhall2020NeRFRS,Chen2024Sketch2NeRFMS,Kim2024GARFieldGA,Song2024Tri2planeVA,Nakayama2024ProvNeRFMP,Ming2024HighQualityMB} include the use of neural networks to represent dynamic human models and the extension of NeRF to dynamic scenes. Additionally, GaussianBody~\cite{Li2024GaussianBodyCH} proposes a segmentation plus scaling strategy to improve point cloud density, and solve the texture blur problem by a pose refinement approach. HDHumans~\cite{Zhou2022HDhumanHH}, jointly optimize neural implicit fields and explicit template meshes, capable of handling loose clothing but with slower rendering speeds. 

\section{Method}    

\subsection{Overview}
In this paper, we aim to efficiently reconstruct high-quality clothed humans from single-view videos involving big movements. Our approach builds on the assumption that the corresponding SMPL parameters $\beta$, $\theta$, the calibrated camera parameters, and the human region mask are all given. Figure~\ref{fig:fig2} illustrates our framework, which takes an initialized point cloud as input and ultimately renders the adjusted Gaussians to produce an image output. We first introduce KGAS which used joint rotations to control the spatial position and orientation during the 3D Gaussian density transformation (Section~\ref{sec:3.2}). The density of the Gaussian distribution is captured by the matrix-Fisher distribution, and the transformation of Gaussians is guided by density. Then, we propose UID  to accurately capture the Gaussian points at corresponding positions based on the direction of the greatest geometric deformation of the reconstructed object's surface. (Section~\ref{sec:3.3}). MOSS differentiates between the human body and its surface, and constrains the newly generated surface Gaussian of the human body using whole-body motion information. This approach allows for the reconstruction of more realistic and accurate details of human surface deformation.

\subsection{Preliminary}
\label{sec:3.1}
 
\subsubsection{SMPL and LBS transformation}
The SMPL human model comprises a human skeleton with $N=6890$ vertices and $k=24$ joints. In this context, let's define the joint rotation as $\theta_{joint} = [\theta_{joint_0}^T,...,\theta_{joint_K}^T]^T$, where $K = 23$, and $\theta_{joint_k}\in R^3$ denotes the rotation of part $k$ relative to its parent in the kinematic tree. Additionally, $\theta_{joint_0}$ refers to the orientation of the root part.

With LBS, the position $p'_i$ of the $i$th vertex $p_i$ in canonical space after skinning is denoted as: 
\begin{equation}
    {p}_i^{\prime}=\sum\nolimits_{k=1}^K w_{k i} G_{0 k}(\theta_{joint}, J(\beta)){p}_i+ G_{1 k}(\theta_{joint}, J(\beta)),
\end{equation}
where $G_{0 k}(\theta_{joint}, J(\beta))$ is the rotation transformation matrix of the joint $k$ that depends on the joint rotation $\theta_{joint}$ and the pose change parameter $\beta$ of human. Similarly, $G_{1 k}(\theta_{joint}, J(\beta))$ is the translation transformation matrix. $w_{k i}$ is the skinning weight of the joint $k$ on vertex $i$, which determines the extent to which joint $k$ affects vertex $i$.

\subsubsection{Matrix-Fisher distribution} 

All 3D rotation matrices lie on a nonlinear and closed manifold SO(3). The matrix-Fisher distribution is defined over SO(3). For a given random rotation matrix $\textbf{R}\in SO(3)$, 3D special orthogonal groups can be defined as $SO(3)=\{\textbf{R}\in\mathbb{R}^{3\times3}|=\textbf{R}^T\textbf{R}=\textbf{I},det(\textbf{R})=1\}$. The matrix-Fisher distribution is obtained according to the matrix rotation parameter denoted as $\theta_{joint_i}$. The probability density function of $R$ is given by the matrix-Fisher distribution:

\begin{equation}
\mathcal{M}(\textbf{R};\theta_{joint_i} ) = p(\textbf{R}|\theta_{joint_i}) = \frac{1}{c(\theta_{joint_i} )} \exp\left(\text{tr}\left( \theta_{joint_i} ^T \textbf{R}\right)\right).
\end{equation}
In this equation, $c(\theta_{joint_i} )$ is a normalization constant that ensures that the total probability sums to 1 on $SO(3)$. We use matrix-Fisher to characterize the joint rotation probability distribution.

The derivation reveals that the most probable rotational state $R_{mode}$ of a given joint (see the Appendix for a detailed derivation) can also be represented by the elements obtained by decomposing $\theta_{joint_i}$ by the proper singular value:

\begin{equation}
R_{mode}={\text{arg max}}\ p\left(R|\theta_{joint_i}\right)=\textbf{U}\textbf{V}^T with\ R \in \text{SO}(3).
\end{equation}

\subsubsection{3D Gaussian Splatting} We use $ f_{\mathrm{3DGS}}=\sum\nolimits_{i}G(x, y, z, \mu_{i}, \Sigma_{i})\cdot c_{i}$ to express 3DGS, where $c_i$ denotes color.
where $G$ is a Gaussian function with inputs as a center (position) $x, y, z$, mean $\mu_i$ and covariance $\sum\nolimits_i$:

\begin{equation}
G(x)=e^{-\frac{1}{2}x^T\boldsymbol{\Sigma}^{-1}x},\boldsymbol{\Sigma}=R_gS_g^TS_gR_g^T.
\end{equation}

\subsection{Kinematic Gaussian Locating Splatting(\textbf{KGAS})}
\label{sec:3.2}

\begin{figure*}[htbp]
    
    \centering
    \includegraphics[width=0.88\linewidth]{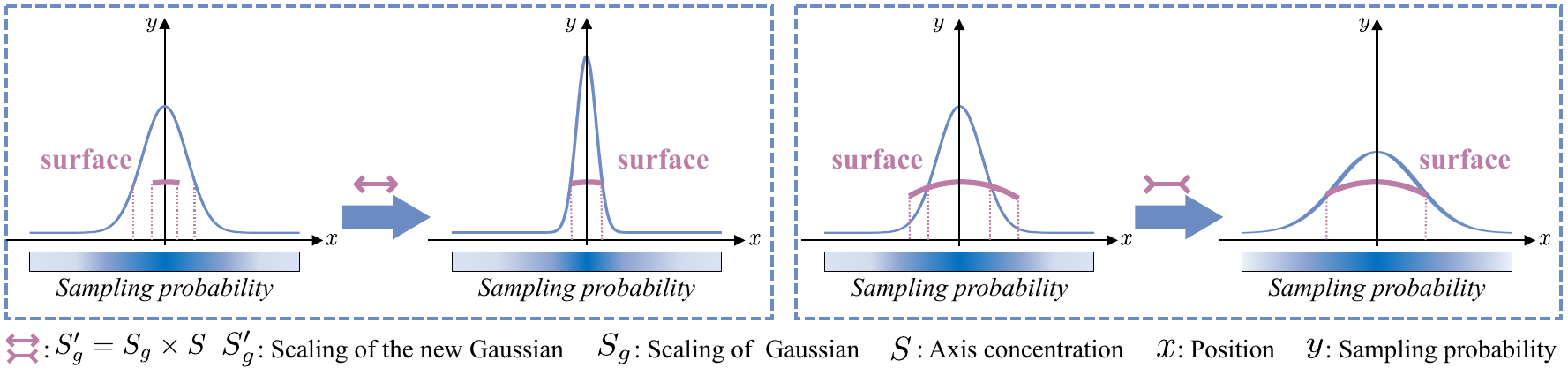}
    \caption{\textbf{Fisher's Gaussian Sampling.} This is a 2D example of how spindle concentration affects Gaussian sampling. note that the color bar represents the probability of sampling, with darker colors representing higher probabilities.}
    \label{fig:fig3}
    
\end{figure*}
 
KGAS performs 3DGS based on global motion information. In this subsection, we will explain how to extract global motion information and how it influences 3DGS, respectively.
 
\subsubsection{Global motion extraction} 
Implementing 3DGS for clothed humans, the Gaussians are typically weakly constrained by the entire body. We attempt to extract global motion and subsequently employ it to guide the 3D Gaussian density transformation. 

To obtain global motion information, we first need to extract local motions. Inspired by~\cite{lee2018bayesian,mohlin2020probabilistic,Sengupta2021HierarchicalKP}, We perform a proper singular value decomposition $\mathbf{\theta_{joint}}$ of the matrix parameters $\mathbf{\theta_{joint}}$ of the joint rotation distribution $\mathbf{\theta_{joint}}=\boldsymbol{U} \boldsymbol{S} \boldsymbol{V}^T$. where $\boldsymbol{U}$ and $\boldsymbol{V}$ are orthogonal matrices, $U$ characterizes the principal axes, and $\boldsymbol{S}=diag(s_1,s_2, s_3)$ represents a diagonal matrix comprising the singular values $s_1>s_2>s_3$ of $\mathbf{\theta_{joint}}$. The singular values dominate the degree of concentration of the distribution along the principal axes, so we use $\boldsymbol{S}$ as evidence to represent the magnitude of directional changes in each joint's movement.

The most probable rotation $R_{mode}$ of the joints is obtained by matrix-Fisher's mode (Section~\ref{sec:3.1}), which contains joint motion information. We utilize $R_{mode}$ to provide directional guidance for the movement of each joint.

Thus, we complete the local human joint motion information extraction. To capture the joint linkage information of the entire body, we introduce the node relationships within the kinematic tree to constrain the matrix-Fisher distribution for each joint. We refer to this approach as Joint-Driven Orientation Refinement (JNTM). We propose to model the matrix-Fisher distribution of the rotation matrices associated with the child nodes of a kinematic tree, given the rotation matrices of their parent nodes. Specifically, we express this as:
\begin{equation}
\mathcal{M}\left(\theta_{joint_i};R_i\right)=p(R_i|\theta_{joint_j},j\in p(i)),
\end{equation}
where $\theta_{joint_i} $ denotes the $i$th joint rotation parameter to be estimated. $\theta_{joint_i} $ can be estimated more exactly by combining the kinematic tree and introducing the prior information of the parent nodes in $p(i)$ by autoregression. The estimated child node $\theta_{joint_i}$ is represented as:

\begin{equation}
\theta_{joint_i} = f_i(\theta_{joint_i},\{\theta_{joint_j}\}_{j\in P(i)}).
\end{equation}
The matrix-Fisher distribution over the SMPL kinematic tree dynamically represents real-time global motion, providing valuable guidance for Gaussian Splatting. Subsequently, we will illustrate how this global motion guides 3D Gaussian Splatting in the following subsection.

\subsubsection{Motion-Based 3D Gaussian Splatting}

When executing large-scale motion, the relative displacement of limbs typically exceeds that of the trunk. However, the Gaussian models that capture cloth folds around the trunk, induced by limb movements, are derived solely from the Gaussians representing the human trunk. Consequently, these models lack crucial global motion information. Therefore, we utilize the global motion extracted in the preceding subsection to guide the density transformation process of the Gaussians, aiming to enhance the representation of the human body during significant motion.

To adjust the size of the sampling area and make it more representative of the actual surface shape, the scaling factor of 3DGS is corrected using the principal axis concentration matrix $\boldsymbol{S}$. We refer to this method as the Density Perceptual Sampler. This contributes to a more accurate representation of the surface deformation by the Gaussians. As shown in  Figure~\ref{fig:fig3}, The corresponding Gaussian distribution $G'(x)$ obtained after the adjustment by $\boldsymbol{S}$ is represented as: 

\begin{equation}
G'(x)=e^{-\frac{1}{2}x^T\boldsymbol{\Sigma}^{-1}x}, 
\end{equation}
where $\boldsymbol{\Sigma}=R_g\boldsymbol{S}^T\boldsymbol{S}R_g^T$. The displacement of the new Gaussian denoted as $\Delta \textbf{x}$ is generated from $G'(x)$, and the position of the center of the new Gaussian ellipsoid $\textbf{x}_{new}=\Delta \textbf{x}+\textbf{x}_{source}$ is obtained from the original Gaussian by this displacement transformation. So the new Gaussian $G_{clone}$ obtained by cloning from the original is represented as:

\begin{equation}
G_{clone}=G(x_{new})=e^{-\frac{1}{2}x_{new}^T\boldsymbol{\Sigma}^{-1}x_{new}}.
\end{equation}
The locations of the resulting new Gaussians are closer to the surface, allowing these Gaussians to more realistically represent the geometric deformation of the surface.

Next, we use the $\boldsymbol{S}$ and $\boldsymbol{R_{mode}}$, which contain global motion information, to control the density transformation process of the Gaussians. The rotation probability concentration matrix $\boldsymbol{S}$ characterizing the different principal axes is used as a scale correction factor for the Gaussian to control the size of the Gaussians. The joint rotation state estimation $\boldsymbol{R_{mode}}$ is used to control the direction of rotation of the Gaussians. matrix-Fisher extracts real-time motion information that is used to guide the Gaussians to learn human posture changes in real-time.
In the Gaussian clone and split process, the estimated $\boldsymbol{S}$ and $\boldsymbol{R_{mode}}$ guide the scaling and rotation of the new Gaussians obtained after cloning, respectively:
\begin{equation}
\boldsymbol{S_g}'=mat\_multi(\boldsymbol{S},\boldsymbol{S_g}), \boldsymbol{R_g}'=mat\_multi(\boldsymbol{R_{mode}},\boldsymbol{R_g}).
\end{equation}
The final Gaussian:
\begin{equation}
G(x_{new})=e^{-\frac{1}{2}x_{new}^T\boldsymbol{\Sigma'}^{-1}x_{new}},   
\end{equation}
where $\boldsymbol{\Sigma'}=\boldsymbol{R'_g}\boldsymbol{S'}^T\boldsymbol{S'}\boldsymbol{R'_g}^T$ are obtained, corresponding to the Gaussian distribution.

Consequently, KGAS achieves Gaussian density perception with the guidance of global motion. Furthermore, the integration of whole-body motion effectively constrains the new Gaussians, thereby enhancing the generation of realistic human body and surface details.

\subsection{Surface Deformation Detector(\textbf{UID})}
 
\label{sec:3.3}
 
\begin{figure*}[htbp]
    
    \centering
    \includegraphics[width=0.88\linewidth]{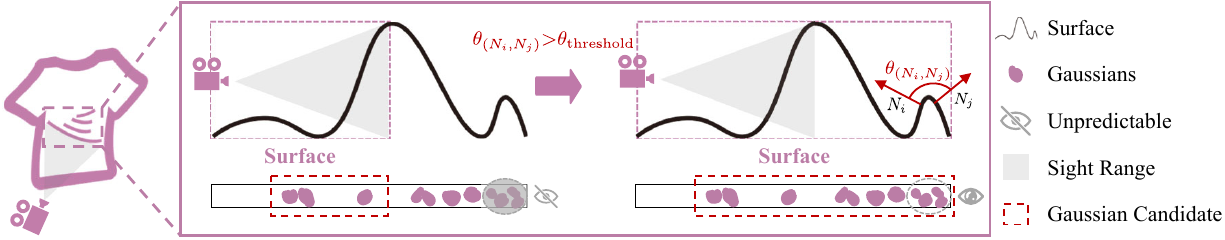}
    \caption{\textbf{Solving occlusion problems with UID (2D).} There is a potential problem of smaller folds being occluded by obvious folds due to the viewing angle. By calculating the degree of directional change in the local distribution of Gaussians, the regions with large deformation on the surface are localized and densely processed.}
    \label{fig:fig4}
    
\end{figure*}

Small folds contribute little to the total pixel-wise loss, leading to weakened supervision. In addition, the mutual occlusion among folds even toughens supervision in single-view. Although~\cite{Kerbl20233DGS,Hu2023GauHumanAG,Li2023Human101T1} pick out Gaussians far from the real human with positional gradients, these are not all for joint details or cloth folds. For this reason, we propose the Gaussian clustering-based UID to detect complex or large surface deformations, see  Figure~\ref{fig:fig4}. 

%
We densify the Gaussians based on gradient localization to increase the rendering granularity for better reproduction of clothing folds and joint details. Remember that the set of Gaussians selected by the gradient is $M=\{m_1,m_2,...\}$. For each point $m_i$ in the point set $M$, select the $k$ Gaussians that are closest to it to form the point set $M_{i_K}$. The local center of mass C of $M_{i_K}$ is given by the following equation:
\begin{equation}
C=\left(\frac{1}{n} \sum\nolimits_{i=1}^n x_i, \frac{1}{n} \sum\nolimits_{i=1}^n y_i, \frac{1}{n} \sum\nolimits_{i=1}^n z_i\right),
\end{equation}
where $n$ is the total number of points in the point set $M_{i_K}$ and $(x_i,y_i,z_i)$ denotes the coordinates of each point.

We compute the covariance matrix of these points in $m_i$ with its neighborhood. Assuming that the local center of mass is $ \bar{m} $ and the points in the neighborhood are $m_{i_k}$, the covariance matrix Cov of the sample is given by the following equation:
\begin{equation}
\operatorname{Cov}=\frac{1}{k-1} \sum\nolimits_{i=1}^k\left(m_{i_k}-\bar{m}\right)\left(m_{i_k}-\bar{m}\right)^T.
\end{equation}
Eigen-decomposition of the covariance matrix is performed to obtain the eigenvalues and eigenvectors:
\begin{equation}
\operatorname{Cov} \cdot v_i=\lambda_i \cdot v_i,
\end{equation}
where $v_i$ is the eigenvector and $\lambda_i$ is the corresponding eigenvalue. The eigenvector corresponding to the smallest eigenvalue of the covariance matrix, $\mathcal{N}_i$, points in the direction of the smallest change in the data~\cite{wang2022deep}. To accurately capture the location of the large deformation, we use the $\mathcal{N}_i$ for each selected Gaussian to calculate the angle $\theta\left(\mathcal{N}_i,\mathcal{N}_j\right)$, which characterizes the magnitude of the localized geometric deformation of the surface.
Specifically, it is computed as follows, given the $\mathcal{N}_i$ of each Gaussian $m_i$ in the set of points with its neighborhood $N_i$ of the other points $m_j$ in the set of points $\mathcal{N}_j$:
\begin{equation}
\theta\left(\mathcal{N}_i,\mathcal{N}_j\right)=arccos\frac{<\mathcal{N}_i,\ \mathcal{N}_j>}{||\mathcal{N}_i||||\mathcal{N}_j||}, j\in N_i.
\end{equation}
If $\theta\left(\mathcal{N}_i,\mathcal{N}_j\right)$ is greater than a threshold, it is considered as a location with large surface deformation, which is labeled and used to guide the generation of more Gaussians at that location to fully learn the human surface deformation.
We precisely capture the hard-to-render surface deformations with folds or protrusions by the minimum direction of geometric deformation and densify them to better reproduce clothing folds and joint details.

\subsection{Loss Function and Training  Optimization}
 
\label{sec:3.4}

Existing methods use random or Structure-from-Motion (SfM)~\cite{snavely2006photo} to initialize 3D Gaussians. We directly sample 6890 Gaussians from the SMPL for initialization to reduce the cost of training. During inference, LBS weights and pose parameters saved in the training phase are used to improve speed while maintaining high rendering quality.

\subsubsection{Loss} Combining image generation quality, human perception, and joint rotation constraints, we adopt the following Loss:
\begin{equation}
L = \lambda_1 L_{image} +\lambda_2L_{percep}+ \lambda_3L_{joint}.
\end{equation}
Among them, $L_{image}$ is the loss used to supervise the quality of image generation, integrating color, and mask. $L_{percep}$ is for human vision perception. $L_{joint}$ constrains the estimation of joint rotational, which enables more accurate parameter estimation for matrix-Fisher distribution (see appendix for specific realizations).

\section{Experiments}

\subsection{Experimental Setups}

\subsubsection{Datasets}

We conduct our experiments on ZJU-MoCap~\cite{NeuralBody_ZJU-MoCap} and MonoCap datasets. ZJU-MoCap includes humans performing complex motions such as licking, punching, etc. It contains foreground masks, camera and SMPL parameters. We pick one camera for training and others for testing. MonoCap dataset is processed by~\cite{Monocap_AS} from DeepCap~\cite{DeepCap} and DynaCap~\cite{DynaCap} dataset. It also provides camera and SMPL parameters as well as human masks. We use one camera view for training and select ten uniformly distributed cameras for testing. 

\subsubsection{Comparison methods}

Compare MOSS with two categories of SOTA methods: Human NeRF-based methods, such as NeuralBody~\cite{NeuralBody_ZJU-MoCap}, HumanNeRF~\cite{HumanNeRF} AnimateNeRF~\cite{AnimateNeRF}, InstantNVR~\cite{InstantNVR} InstantAvatar~\cite{InstantAvatar}. 3DGS methods, including GauHuman~\cite{Hu2023GauHumanAG} and Human101~\cite{Li2023Human101T1}. Following~\cite{InstantNVR}, for a fair comparison, we do not optimize the test time for the SMPL parameters using images from the test set on InstantAvatar~\cite{InstantAvatar}.

\subsubsection{Metrics}
PSNR, SSIM, and LPIPS* are utilized as indicators of quality, while FPS is employed as a metric for speed. PSNR evaluates the variance between pixel values of two images. SSIM measures image similarity in terms of luminance, contrast, and structural integrity. LPIPS* compares two images in the feature space, which is considered to be closer to human visual perception.

\subsubsection{Implementation Details}
We utilize AdamW with varying parameters across different modules. Experiments were carried out on a single RTX 3090 GPU, with the model taking approximately 3,000 iterations to converge. Additionally, a detailed overview of our network structure and hyperparameters can be found in the appendix.

\subsection{Evaluation}
\label{sec:4.2}
\begin{table*}[ht]
\caption{Quantitative evaluation of our method compared with previous works. We conduct experiments on the ZJU-MoCap and MonoCap datasets and list the average of metrics at a resolution of 512 $\times$ 512.}
\centering
\resizebox{0.7\textwidth}{!}{
\begin{tabular}{l|c c c c | c c c c c c}
\toprule
\multirow{2}{*}{Method}&\multicolumn{4}{c}{ZJU-MoCap}&\multicolumn{4}{c}{ MonoCap}\cr  
			\cmidrule(lr){2-5} \cmidrule(lr){6-9}

              &PSNR↑     & SSIM ↑   & LPIPS*↓            & FPS        & PSNR↑  & SSIM ↑ & LPIPS*↓     & FPS     \\
\midrule
NeuralBody    & 29.03     & 0.964    & 42.47                    & 1.48       & 32.36  & 0.986  & 16.70              & 0.98    \\
HumanNeRF     & 30.66     & 0.969    & 33.38                    & 0.30        & 32.68  & 0.987  & 15.52            & 0.08    \\
AnimateNeRF   & 29.77     & 0.965    & 46.89                   & 1.11       & 31.07  & 0.985  & 19.47             & 0.31    \\
InstantNVR    & 31.01     & \underline{0.971} & 38.45              & 2.20        & 32.61  &\underline{0.988} & 16.68       & 0.75    \\
InstantAvatar & 29.73     & 0.938    & 68.41                     & 4.15       & 30.79  & 0.964  & 39.75              & 2.54    \\
Human101     & \underline{31.79} & 0.965 & 35.75                 & 104        & 32.63  & 0.982  & 16.51             & 104     \\
GauHuman      & 31.34     & 0.965    & 30.51                    & 189        & 33.45  & 0.985  & 13.35             & 189     \\
\textbf{MOSS} (Ours) & 31.01 & 0.963 & \underline{25.40}                    & \underline{276}               & \underline{33.56}              & 0.985                & \underline{11.68}                             & \underline{257}                   \\
\bottomrule
\end{tabular}
}
\label{tab:tab1}

\end{table*}

To validate the effectiveness of MOSS in solving the problem of lacking global constraints for clothed human reconstruction, we compare our method with some previous human reconstruction methods~\cite{NeuralBody_ZJU-MoCap, HumanNeRF, AnimateNeRF, InstantNVR, InstantAvatar, Hu2023GauHumanAG, Li2023Human101T1} on the ZJU-MoCap and MonoCap datasets. Our method outperforms the comparison methods on most of the metrics, as shown in Table~\ref{tab:tab1}, thus validating the effectiveness of MOSS in solving the problem. Detailedly, MOSS achieves state-of-the-art LPIPS* on ZJU-MoCap and MonoCap, and the PSNR also achieves state-of-the-art on the Monocap dataset. It quantitatively demonstrates that our approach can maintain excellent rendering results while achieving industry-leading FPS.

GauHuman initializes with SMPL and prunes the 3D Gaussians with large KL divergence in adaptive density control, which only provides static human body information. On the contrary, we propose KGAS to control the Gaussian Splatting through the density and rotation factors. Compared with GauHuman, our method improves the LPIPS* by 16.75\% and 12.51\% on the ZJU-MoCap and MonoCap datasets, respectively, and improves the PSNR by 0.11 on the MonoCap dataset, which illustrates the effectiveness of our method.

Compared with InstantNVR, our method improves the LPIPS* on ZJU-MoCap and MonoCap datasets by 33.94\% and 29.98\%, respectively. The PSNR on the MonoCap dataset improves by 0.95. InstantNVR manually divides the human body into multiple parts based on SMPL, mapping the vertices to the UV coordinate and the in-depth learning of each part. Although this effectively enhances the structural information of the image, it still lacks objectivity due to the manual division and the human body cannot be divided finely as well. On the contrary, the UID we designed based on KGAS can detect the local position with large deformation by objectively searching the surface and strengthening the supervision. 

\begin{figure}[h]
    
    \centering
    \includegraphics[width=0.95\linewidth]{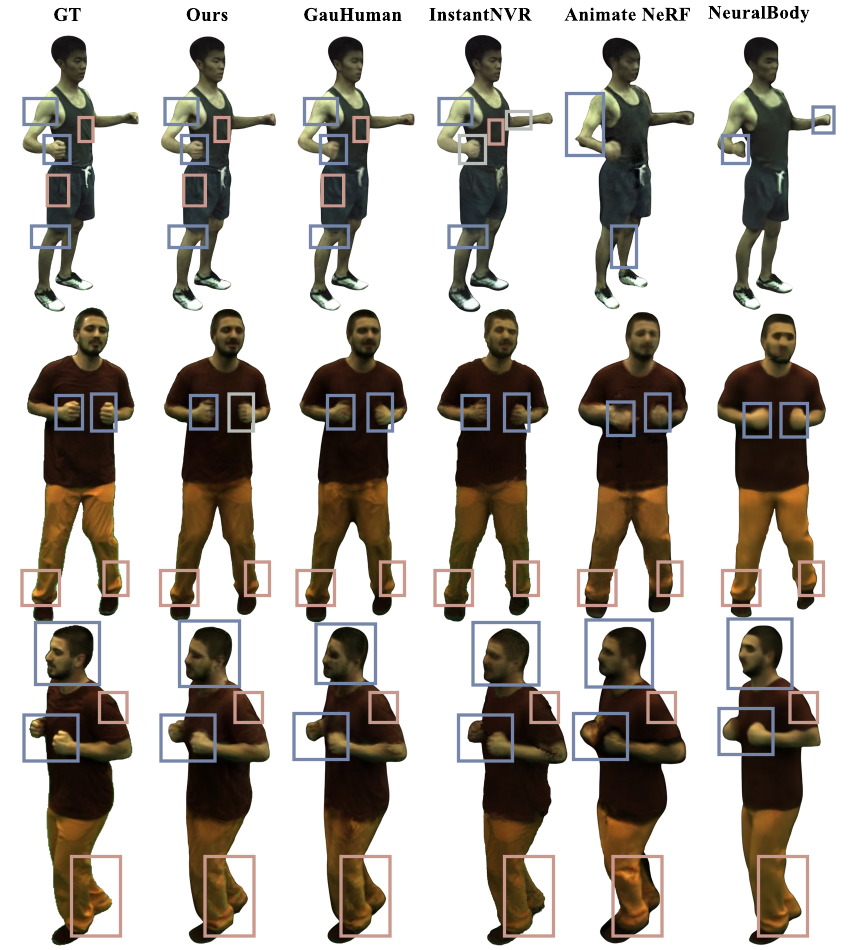}
    \caption{To ensure a fair comparison, we compare~\cite{NeuralBody_ZJU-MoCap,HumanNeRF,AnimateNeRF,InstantNVR,Hu2023GauHumanAG,Li2023Human101T1} at 512 $\times$ 512 resolution. Our model shows better visual quality and more detail.}
    \label{fig:fig5}
    
\end{figure}

\subsubsection{Reconstruction Efficacy of MOSS} 

We validate the constraint effect of MOSS on 3D Gaussians, as well as its ability to identify local regions with significant surface deformations. As shown in  Figure~\ref{fig:fig5}, we select the same human body from two different viewpoints and the qualitative comparisons. The details of the human body joints are marked by the red box, and the details of the clothing folds and other deformations are marked by the yellow box. It shows that our method can reproduce the surface details and clothing folds with high quality in large motion scenarios.

\subsubsection{Gaussian Guidance}
As shown in  Figure~\ref{fig:fig5}, MOSS achieves more excellent visual quality in the details of human joints. The edges of clothing and joints are clearly demarcated, and the \textbf{orientation} of clothing folds is closer to the real occasion (e.g., shorts and thighs). Among the compared methods, NeuralBody encodes human body structure with the latent code of SMPL vertices, but it cannot capture complex human movement. This results in blurred fists and incomplete hands in the figure. AnimateNeRF expects to overcome the ill-posed problem. The human joint skeleton is introduced as regularisation to facilitate training. Its rendering results still have artifacts in arms and calves. GauHuman's Gaussians lack global constraints, leading to problem such as confusion between shorts and knees, as well as loss of hand details. The KGAS module of MOSS makes the human body reconstruction constrained by the topology of the kinematic tree and guides the 3D Gaussian displacements with the global motion, which is able to fully learn the \textbf{motion of human}, and the \textbf{differences between the clothes and the human body}. In addition, the orientation of clothing folds is also more realistic due to the constraints of motion direction.

\subsubsection{Surface Deformation Detection}
MOSS restores more surface details. On the contrary, InstantNVR's modeling only considers the human body and ignores the clothing, resulting in a complete loss of clothing folds of the human reconstructed by InstantNVR. What's more, the InstantNVR rendered shorts tightly clinging to the legs, which is different from the loose shorts in reality. The pixel-based loss used in GauHuman only provides weak supervision, which results in its restoration of only a few folds (e.g., the sides of the waistcoat and shorts on the torso shown in the yellow box in the first row of results). The UID \textbf{detects complex deformations} on the local body surface and compensates for the geometric reconstruction. Hence, UID effectively \textbf{solves the problem of insufficient supervision in a single-view}, and the folds that are hidden from each other can also be restored.

\begin{figure*}[htbp]
    
    \centering
    \includegraphics[width=0.85\linewidth]{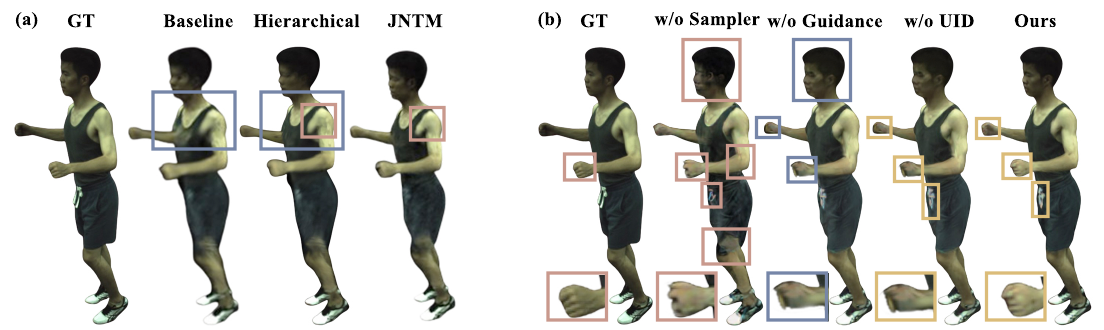} 
    
    \caption{\textbf{Ablation studies result of rendered images.} (a) Results of the JNTM are demonstrated. (b) Reflecting the validity by removing our proposed components.}
    \label{fig:fig6}
    
\end{figure*}

\begin{figure*}[htbp]
    \centering
    \includegraphics[width=0.85\linewidth]{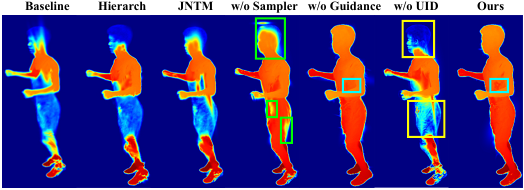} 
    
    \caption{\textbf{Ablation studies result of depth images.} Warmer tones indicate denser information and cooler tones indicate sparser information.}

    \label{fig:fig7}
    
\end{figure*}

\begin{table}[h]
	\centering  
 \caption{The effectiveness of Joint-Driven Orientation Refinement and its components}
	\fontsize{8}{15}\selectfont  
		\label{tab:tab2}
  \resizebox{1\linewidth}{!}{
		\begin{tabular}{l|c c c}
			\toprule  
            Method         & LPIPS*↓ & PSNR↑ & SSIM ↑ \\
            \midrule  
            Baseline       & 37.62   & 28.91 & 0.963  \\
            Hierarchical (Baseline + Autoregression) \  \  \ \ \ & 32.55   & 30.12 & 0.971  \\
            JNTM (Hierarchical +  joint linkage)           & 24.98   & 30.87 & 0.971  \\
    \textbf{Ours}&\underline{15.18}&\underline{31.84}&\underline{0.974}\\
    \bottomrule
\end{tabular}  
  }

\label{tab:tab2}

\end{table}

\begin{table}[h]
\centering  
\fontsize{8}{15}\selectfont  
\caption{Evaluations to ensure the functionality of each component of MOSS.}
    \label{tab:tab3}
    \resizebox{1\linewidth}{!}{
     \begin{tabular}{l|c c c}
        \toprule  
            Method                         & LPIPS*↓ & PSNR↑ & SSIM ↑  \\
        \midrule  
            \textbf{Ours}&\underline{15.18}&\underline{31.84}&\underline{0.974}\\
            w/o Density Perceptual Sampler \ \ \ \ \ \ \ \ \ \ \ \ \ \ \ \ & 18.53   & 31.31 & 0.97    \\
            w/o Motion Guidance            & 16.89   & 31.75 & 0.973   \\
            w/o UID                        & 20.71   & 31.07 & 0.968   \\
    \bottomrule
\end{tabular}   
    }

\label{tab:tab3}

\end{table}

\subsection{Ablation Study}
We conduct ablation studies to examine the role that every component plays in MOSS. The results of our experiments on ZJU-MoCap are shown in Table~\ref{tab:tab2}, Table~\ref{tab:tab3},  Figure~\ref{fig:fig6} and  Figure~\ref{fig:fig7}. Table~\ref{tab:tab2} presents the effectiveness of JNTM and its sub-components. The baseline is 3D Gaussian Splatting initialized with SMPL. By adding Autoregression and  joint linkage \textbf{incrementally}, the quality of human body reconstruction is progressively improved. Table~\ref{tab:tab3} shows the results of MOSS and the outcomes after removing only \textbf{one component each time}. Through selectively disabling each component, we evaluate their contribution to the system. We will analyze the efficacy of individual components according to experimental results specifically in the following part.

\subsubsection{Joint-Driven Orientation Refinement}
As is shown in Table~\ref{tab:tab2}, the \textbf{Autoregression} improves the performance in LPIPS*, PSNR, and SSIM. This validates that it is effective to use the kinematic tree to constrain the parameter estimation for the distribution of joint rotations. Furthermore,  joint linkage also improves the metrics, this is because it precisely controls Gaussians by joint movement. In comparison, Ours has significantly enhanced the LPIPS*, owing to the incorporation of guidance on Gaussians and surface deformation detection. The rendered and depth images in  Figure~\ref{fig:fig6} (a) and  Figure~\ref{fig:fig7} also show the progressively improved effect when sequentially adding subcomponents of JNTM.

\subsubsection{Density Perceptual Sampler}
We conduct the following ablation experiments to explore the Gaussian density and surface structure perceptual ability. The results are shown in Table~\ref{tab:tab3} w/o Density Perceptual Sampler. According to the green box in  Figure~\ref{fig:fig6} (b), the absence of this component results in the blurred rendering of surface edges, especially the fingers and the face. It indicates that the Gaussians of the human body and surface are severely underdense on the boundaries in  Figure~\ref{fig:fig7}. In comparison, the Density Perceptual Sampler uses \textbf{the principal axis concentration} to correct the sampling probability of the Gaussian split process, such that the \textbf{sampling area is more aligned with the actual surface}. Besides, the restored deformation is more significant and clear even if the human is performing motions such as head rotation, finger clenching, lumbar rotation, and foot padding.

\subsubsection{Motion Guidance}
The experimental results highlight the impact of Motion Guidance. Human body movements, such as fist clenching or arm raising, lead to local shadows. Comparison of blue boxes in  Figure~\ref{fig:fig6} (b) and  Figure~\ref{fig:fig7} shows that Motion Guidance can mitigate the confusion between shadows and actual body parts. Additionally, the blue circle illustrates that Motion Guidance helps to restore the precise shapes of irregular and flexible clothing (e.g., belt loop). This component guides Gaussians with global motion, dynamically adjusting the \textbf{orientation and scaling} of each Gaussian. Consequently, it not only enhances the fidelity of reproducing local shadows in human body motion but also improves the accuracy in describing texture and shape alterations in clothing. The efficacy of w/o Motion Guidance is further elucidated through the results of ablation experiments in Table~\ref{tab:tab3}.

\subsubsection{UID}
To verify UID's ability to identify the complex folds on the surface, we perform the following ablation experiments, with the results shown in  Figure~\ref{fig:fig6} (b). If we omit UID and only use pixel-wise losses for supervision, it will lead to insufficient guidance. Especially in  Figure~\ref{fig:fig7}, most of the body parts lack depth information. Hence, it gets more difficult to accurately reconstruct various human body surfaces (especially under occlusion scenarios, seen in  Figure~\ref{fig:fig4}). Furthermore, a comparison between the metrics of w/o UID and Full in Table~\ref{tab:tab3} reveals a decline in the performance across all evaluation metrics. Confusion between the human body and trunk clothing (see yellow boxes), surface rendering distortions (see the yellow circle), and significant loss of facial features, including drastic texture changes and loss of information in high-frequency regions (see the yellow polygon), are evident. In addition,  Figure~\ref{fig:fig6} also shows that without UID leads to inaccurate restoration of overlapping lower legs, knees, and shorts with significant artifacts.

\section{Conclusion}

To address the lack of global constraints for detailed reconstruction of clothed humans in motion, we propose MOSS for reconstructing humans from 3DGS guided by global motion. The framework sends the human body in motion prior to the Gaussian rendering process of the body surface and focuses on the locations where the surface deformation is significant. In our future work, we consider incorporating the graph to topologically guide 3D clothed human reconstruction. There is also a wide range of substantial human motion scenarios in many fields, such as virtual reality and the fashion industry, where our technology has potential applications. For instance, it could reduce game production costs, enhance player experience, and support fashion designers in optimizing their designs.

\bibliographystyle{ieeenat_fullname}
\bibliography{main}

\clearpage
\appendix
\vspace*{1em}{\centering\Large\bf%
Appendix
\vspace*{1.5em}}

In this Supplementary Material, we will further detail the following aspects omitted in the main text.
\begin{enumerate}
\item Section \ref{Sec. A} presents the detailed mathematical derivation regarding the matrix-Fisher distribution.
\item Section \ref{Sec. B} provides the implementation details of MOSS and the Loss we used in the main text.
\item Section \ref{Sec. C} analyzes metrics used for qualitative evaluation.
\item Section \ref{Sec. D} shows more experiments and more results.
\end{enumerate}

\section{Mathematical Derivation} 
\label{Sec. A}

\subsection{Proper Singular Value Decomposition.}

In this paper, we use proper singular value decomposition for joint rotation $\theta_{joint}$. Here, we are going to explain why it is necessary to use \textbf{proper singular value decomposition} instead of the common \textbf{singular value decomposition}.
Suppose the singular value decomposition of $\theta_{joint}$ is given by
\begin{equation}
\label{eq 1}
\theta_{joint}=U^{\prime} S^{\prime}\left(V^{\prime}\right)^T,
\end{equation}

where $S^{\prime} \in \mathbb{R}^{3 \times 3}$ is a diagonal matrix composed of the singular values $s_1^{\prime} \geq s_2^{\prime} \geq s_3^{\prime} \geq 0$ of $F$, sorted in descending order, and $U^{\prime}, V^{\prime} \in \mathbb{R}^{3 \times 3}$ are orthonormal matrices. Since $\left(U^{\prime}\right)^T U^{\prime}=\left(V^{\prime}\right)^T V^{\prime}=I_{3 \times 3}$ the determinant of $U^{\prime}$ or $V^{\prime}$ is $\pm 1$. Note the orthogonal matrices $U^{\prime}$ and $V^{\prime}$ are not necessarily rotation matrices in $\mathrm{SO}(3)$, as their determinant is possibly -1 . To resolve this, we introduce a proper singular value decomposition as follows.
For a given $\theta_{joint}\in \mathbb{R}^{3 \times 3}$ , let the singular value decomposition be given by Equation \ref{eq 1}. The \textbf{proper} singular value decomposition of $\theta_{joint}$ is defined as
\begin{equation}
\label{eq 2}
\theta_{joint}=U S V^T,
\end{equation}
where the rotation matrices $U, V \in \mathrm{SO}(3)$. The diagonal matrix $S \in \mathbb{R}^{3 \times 3}$ are
\begin{align}
    U & =U^{\prime} \operatorname{diag}\left[1,1, \operatorname{det}\left[U^{\prime}\right]\right], \\
S & =\operatorname{diag}\left[s_1, s_2, s_3\right]=\operatorname{diag}\left[s_1^{\prime}, s_2^{\prime}, \operatorname{det}\left[U^{\prime} V^{\prime}\right] s_3^{\prime}\right], \\
V & =V^{\prime} \operatorname{diag}\left[1,1, \operatorname{det}\left[V^{\prime}\right]\right] .
\end{align}

The definitions of $U$ and $V$ are formulated such that $\operatorname{det}[U]=\operatorname{det}[V]=+1$ to ensure $U, V \in \mathrm{SO}(3)$. Note that the first two proper singular values $s_1, s_2$ are non-negative, but the last one $s_3$ could be negative when $\operatorname{det}\left[U^{\prime} V^{\prime}\right]=-1$ and $s_3^{\prime}>0$. As $s_1^{\prime}, s_2^{\prime}, s_3^{\prime}$ are sorted in descending order,
\begin{align}
0 \leq\left|s_3\right| \leq s_2 \leq s_1, \\
0 \leq s_2+s_3 \leq s_3+s_1 \leq s_1+s_2 .
\end{align}

\subsection{Proof for the Mode of the Matirx-Fisher Distribution.}
 Let the probability density of $R \in \mathrm{SO}(3)$ be $p(R): \mathrm{SO}(3) \rightarrow \mathbb{R}$. Suppose $R \sim \mathcal{M}(F)$ with a matrix $\theta_{joint} \in \mathbb{R}^{3 \times 3}$. Let the proper singular value decomposition of $\theta_{joint}$ be given by Equation \ref{eq 2}. Then, the first moment of $R$ is given by
 \begin{equation}
 \label{eq 8}
     \mathrm{E}[R]=U \mathrm{E}[Q] V^T.
 \end{equation}

The max mean is defined as
\begin{equation}
\label{eq 9}
    \mathrm{M}_{\max }[R]=\underset{R \in \operatorname{SO}(3)}{\arg \max }\{p(R)\},
\end{equation}

and the minimum mean square error (MMSE) mean is defined as
\begin{equation}
\label{eq 10}
    \mathrm{M}_{\mathrm{mse}}[R]=\underset{R \in \mathrm{SO}(3)}{\arg \min }\left\{\int_{\mathrm{SO}(3)}\|R-\tilde{R}\|_F^2 p(\tilde{R}) d \tilde{R}\right\} .
\end{equation}

For the matrix Fisher distribution, the above most probable joint rotations are equivalent and they can be obtained explicitly as follows.
Next, from the definition of the Frobenius norm
\begin{equation}
\label{eq 11}
    \|R-\tilde{R}\|_F^2=\operatorname{tr}\left[(R-\tilde{R})^T(R-\tilde{R})\right]=\operatorname{tr}\left[2\left(I_{3 \times 3}-R^T \tilde{R}\right)\right] .
\end{equation}

Substituting Equation \ref{eq 11} into Equation \ref{eq 9}, and we use $\int_{\mathrm{SO}(3)} p(\tilde{R}) d \tilde{R}=1$ and $\int_{\mathrm{SO}(3)} \tilde{R} p(\tilde{R}) d \tilde{R}=\mathrm{E}[R]$, we can derive that
\begin{equation}
    \mathrm{M}_{\mathrm{mse}}[R]=\underset{R \in \mathrm{SO}(3)}{\arg \min }\left\{6-2 \operatorname{tr}\left[R^T \mathrm{E}[R]\right]\right\} .
\end{equation}

Substituting Equation \ref{eq 8}, it is straightforward to show $\mathrm{M}_{\mathrm{mse}}$ is the value of $R$ that maximizes $\operatorname{tr}\left[R^T \mathrm{E}[R]\right]=\operatorname{tr}\left[R^T U \mathrm{E}[Q] V^T\right]=$ $\operatorname{tr}\left[\mathrm{E}[Q]\left(V^T R^T U\right)\right].$
According to ~\cite{lee2018bayesian}, the diagonal elements of $\mathrm{E}[Q]$ serve as the proper singular values, and the probability density of a matrix Fisher distribution is maximized when $V^T R^T U=U^T R V=I_{3 \times 3}$, or $R=U V^T$, which is the max most probable joint rotation. 
That is to say, the max mean and the minimum mean square mean of $R$ are identical and they are given by
\begin{equation}
\label{eq 13}
    \mathrm{M}_{\text {max }}[R]=\mathrm{M}_{\text {mse }}[R]=U V^T \in \mathrm{SO}(3) .
\end{equation}

What's more, the MSE expressed by Equation \ref{eq 10} can serve as s Loss when estimating $R=U V^T$, for more details of Loss please see in Section \ref{Sec. B})

\subsection{Geometric Interpretation of the Matirx-Fisher Distribution.}
Next, we provide the geometric interpretation of the matrix parameter $F$ in determining the shape of the matrix Fisher distribution according to ~\cite{lee2018bayesian}. Consider a set of rotation matrices parameterized by $\theta_i \in[0,2 \pi)$ for $i \in\{1,2,3\}$ as
\begin{equation}
\label{eq 14}
    R_i\left(\theta_i\right)=\exp \left(\theta_i \widehat{U e_i}\right) U V^T=U \exp \left(\theta_i \hat{e}_i\right) V^T .
\end{equation}

This corresponds to the rotation of the most probable joint rotation $U V^T$ of $\mathcal{M}(F)$, about the axis $U e_i$ by the angle $\theta_i$, where $U e_i$ is considered expressed with respect to the inertial frame.
Using Equation \ref{eq 13}, the probability density along Equation \ref{eq 14} is given by
\begin{equation}
\label{eq 15}
    p\left(R_i\left(\theta_i\right)\right)=\frac{1}{c(S)} \exp \left(\operatorname{tr}\left[S \exp \left(\theta_i \hat{e}_i\right)\right]\right) .
\end{equation}

Substituting Rodrigues' formula ~\cite{shuster1993survey}, namely  $\exp \left(\theta_i \hat{e}_i\right)=I_{3 \times 3}+\sin \theta_i \hat{e}_i+\left(1-\cos \theta_i\right) \hat{e}_i^2$, and rearranging, this reduces to
\begin{equation}
\label{eq 16}
    p\left(R_i\left(\theta_i\right)\right)=\frac{e^{s_i}}{c(S)} \exp \left(\left(s_j+s_k\right) \cos \theta_i\right),
\end{equation}
where $j, k$ are determined such that $(i, j, k) \in \mathcal{I}$. This resembles the von Mises distribution on a circle ~\cite{mardia2009directional}, where the probability density is proportional to $\exp ^{\kappa \theta}$ with a concentration parameter $\kappa=s_j+s_k$. As such, when considered as a function of $\theta_i$, the distribution of $p\left(R_i\left(\theta_i\right)\right)$ becomes a uniform distribution as $s_j+s_k \rightarrow 0$, and it is more concentrated as $s_j+s_k$ increases. It has been also shown that when $s_j+s_k$ is sufficiently large, the von Mises distribution of $\theta_i$ is well approximated by the Gaussian distribution with the variance of $\frac{1}{s_j+s_k}$ ~\cite{mardia2009directional}.

Another noticeable property of Equation \ref{eq 16} is that the probability density depends only on the singular values $s_i, s_j, s_k$ and the rotation angle $\theta_i$, and it is independent of $U$ or $V$. Recall [43] corresponds to the rotation of the most probable joint rotation $U V^T$ about the $i$-th column of $U$. Consequently, each column of $U$ is considered as the principal axis of rotation for $\mathcal{M}(\theta_{joint})$, analogous to the principal axes of a multivariate Gaussian distribution.

In summary, the role of $F=U S V^T$ in determining the shape of the distribution of $\mathcal{M}(\theta_{joint})$ is as follows: 

\begin{enumerate}[label=(\roman*)]
\item The \textbf{most probable joint rotation} is given by $U V^T$ (denoted as $R_{mode}$). 

\item The columns of the rotation matrix $U$ specify the principle axes of rotations in
the inertial frame.

\item The proper singular values $S$ describe \textbf{the concentration of the distribution along the rotations about the principle axes}, and in particular, the dispersion along the rotation of the most probable joint rotation about the axis $U e_i$ is determined by $s_j+s_k$ for $(i, j, k) \in \mathcal{I}$.
\end{enumerate}

For instance, consider $\theta_{joint}=\operatorname{diag}[25,5,1]$, where $\left(s_1, s_2, s_3\right)=(25,5,1)$ and $U=V=I_{3 \times 3}$ . The most probable joint rotation is $I_{3 \times 3}$, and the principal axes are $\left(e_1, e_2, e_3\right)$. Since $s_2+s_3=6 \leq s_3+s_1=26 \leq s_1+s_2=30$, rotating the most probable joint rotation about the first principal axis $e_1$ (lower left) is most dispersed, thereby making the marginal distribution of the second and the third body-fixed axes elongated along the great circle perpendicular to the first principal axis.

\begin{figure}[h]
    \centering
    \includegraphics[width=1\linewidth]{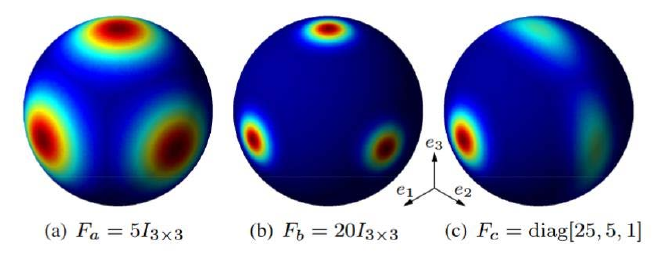}
    \caption{ Visualization of selected matrix Fisher distributions from ~\cite{lee2018bayesian} .the distribution in (b) is more concentrated than in (a), as the singular values of $F_b$ are greater than those of Fa; for both (a) and (b), the distributions of each axis are identical and circular as three singular values of each of 
$F_a$ and $F_b$ are identical; in (c), the first body-fixed axis (lower left) is more concentrated as the first singular value of  $F_c$  is the greatest, and the distributions for the other two axes are elongated.}
    \label{fig:fig1}
\end{figure}

\section{Implemention Details}
\label{Sec. B}
\subsection{Implementation Details of MOSS. }
Autoregression consists of 4 fully connected layers, i.e., an input layer, 2 hidden layers, and one output layer. Each layer is followed by a ReLU activation. The dimension of the hidden layer is 128, while the input and output dimensions of the pose refinement module are 69. These layers are utilized to exact the features of nodes. The parent nodes' information will be added to the child node when the last layer has the exact feature. Subsequently, it employs a configuration of 5 fully connected layers, encompassing one input layer, 3hidden layers, and a single output layer, to determine the LBS offset. The ReLU activation is used after each layer. Positional encoding is applied to the 3D Gaussian positions before they are fed into the input layer. The input and output dimensions of the LBS offset module are 63 and 24 (number of joints). To enhance the understanding of the spatial connections between different body parts facilitated by bone rotations, the proposed method incorporates an attention module. Takes $\theta_{joint}$ and $w_{bise}$ as inputs, obtains $\mathrm{Q}, \mathrm{K}, \mathrm{V}$ through the corresponding linear projection layers $\phi$, and calculates the matching score:
\begin{align}
 \boldsymbol{Q}&=\phi_q(\boldsymbol{w_{bise}}), 
 \boldsymbol{K}=\phi_k(\boldsymbol{\theta_{joint}}), 
 \boldsymbol{V} =\phi_v(\boldsymbol{\theta_{joint}}),\\
\quad\boldsymbol{A}&=\operatorname{softmax}\left(\frac{\boldsymbol{Q} \boldsymbol{K}^T}{\ sqrt(d)}\right)V,
\end{align}
where $d$ is the dimension of $\boldsymbol{Q}$ used to scale the dot product to prevent the gradient from vanishing, $\boldsymbol{A} \in \mathbb{R}^{N \times 24}$. We use AdamW optimizer with a learning rate $2.5 \times 10^{−4}$ and $1 \times 10^{-4}$ to optimize the above two modules. Other training details for 3D Gaussians are the same as ~\cite{Kerbl20233DGS}.

\subsection{Efficiency Enhancement.}

As the covariance matrix is decomposed into the product of rotation and scaling matrices $\boldsymbol{\Sigma}=\boldsymbol{R} \boldsymbol{S} \boldsymbol{S}^T \boldsymbol{R}^T$. $S$ is a diagonal matrix, the inverse and determinant of a diagonal matrix can be easily derived. $R$ is an orthogonal matrix that enjoys the following properties:
$\boldsymbol{R}  \boldsymbol{R}^T=I.$
Thus, it is more efficient to compute the following matrix in optimization:
\begin{align}
\boldsymbol{\Sigma}^{-1} & =\left(\boldsymbol{R} \boldsymbol{S} \boldsymbol{S}^T \boldsymbol{R}^T\right)^{-1}=\boldsymbol{R} \boldsymbol{S}^{-1} \boldsymbol{S}^{-1} \boldsymbol{R}^T, \\
\operatorname{det} \boldsymbol{\Sigma} & =\operatorname{det}\left(\boldsymbol{R} \boldsymbol{S} \boldsymbol{S}^T \boldsymbol{R}^T\right)=\operatorname{det}(\boldsymbol{S}) * \operatorname{det}(\boldsymbol{S}).
\end{align}
 
\subsection{Details of Loss Functions.} 
The Loss used in this paper combines image generation quality, human perception, and joint rotation. Among them, $L_{image}$ is the loss used to supervise the quality of image generation, integrating color, and mask. $L_{percep}$ is the loss of human image perception, including random structural similarity and LPIPS. $L_{joint}$ is the joint that constrains the estimation of the joint rotational state, which enables more accurate parameter estimation for Fisher distribution. The detailed loss computation and the weight for each loss function are as follows.

\subsubsection{Image Loss.} Utilize the $L_2$ normalization to constrain the predicted result:
\begin{equation}
L_{image} = L_{color} + \alpha_1L_{mask},
\end{equation}
where $\alpha_1$ is set 0.5. More specifically:
Photometric Loss. Given the ground truth target image $C$ and predicted image $\hat{C}$, we apply photometric loss:
\begin{equation}
L_{\text {color }}=\|\hat{C}-C\|_2 .
\end{equation}

Mask Loss. We also leverage the human region masks for optimization. The mask loss is defined as:
\begin{equation}
L_{\text {mask }}=\|\hat{M}-M\|_2,
\end{equation}

where $\hat{M}$ is the accumulated volume density and $M$ is the ground truth binary mask label.

\subsubsection{Perception Loss.}We further employ SSIM and S3IM to ensure the structural similarity between ground truth and synthesized images. The perceptual loss LPIPS is also utilized to ensure the quality of rendered image, i.e.:

\begin{equation}
L_{percep}=\alpha_2\operatorname{SSIM}(\hat{C}, C) + \alpha_3\operatorname{S3IM}(\hat{C}, C) + \alpha_4\operatorname{LPIPS}(\hat{C}, C) .
\end{equation}

Where $\alpha_2,\alpha_3$, and $\alpha_4$, which are set to 0.2, 0.5, and 0.3 respectively. 
SSIM Loss. We further employ SSIM to ensure the structural similarity between ground truth and synthesized images:

\begin{equation}
L_{S S I M}=1-\operatorname{SSIM}(\hat{C}, C) .
\end{equation}

\subsubsection{S3IM Loss.}  Recently proposed by ~\cite{xie2023s3im}, it introduces stochastic structural similarity, using the similarity between two sets of pixels as the training loss. Each pixel set generally contains thousands of pixels, which together contribute to correlated, global structural information.
\begin{equation}
\operatorname{S3IM}(\hat{\mathcal{R}}, \mathcal{R})=\frac{1}{M} \sum_{m=1}^M \operatorname{SSIM}\left(\mathcal{P}^{(m)}(\hat{\mathcal{C}}), \mathcal{P}^{(m)}(\mathcal{C})\right),
\end{equation}
where SSIM needs to apply the kernel size $K \times K$ and the stride size $s$.
We note that computing S3IM can be well vectorized and multiplex training also only requires back-propagation for once per iteration. Thus, the extra computational cost of multiplex training is limited. As S3IM lies in $[-1,1]$ and positively correlated with image quality, we define the S3IM-based loss $L_{\mathrm{S} 3 \mathrm{IM}}$ as:
\begin{align}
L_{\mathrm{S} 3 \mathrm{IM}}(\Theta, \mathcal{R}) & =1-\operatorname{S} 3 \operatorname{IM}(\hat{\mathcal{R}}, \mathcal{R}) \\
& =1-\frac{1}{M} \sum_{m=1}^M \operatorname{SSIM}\left(\mathcal{P}^{(m)}(\hat{\mathcal{C}}), \mathcal{P}^{(m)}(\mathcal{C})\right).
\end{align}

\subsubsection{LPIPS Loss.} The perceptual loss LPIPS is also utilized to ensure the quality of rendered images as:
\begin{align}
L_{\text {LPIPS }}=\operatorname{LPIPS}(\hat{C}, C) .
\end{align}

\subsubsection{Joint Loss.}$L_{joint}$ is just the negative log-likelihood (NLL) loss on distribution parameters to constrain the matric-Fisher distribution, which is proved in Section \ref{Sec. A}. Notably, this loss function is assigned 0.06.
For each joint rotation matrix $R_i$ we model it with a matrix-Fisher distribution $\mathcal{M}(\boldsymbol{R_i} ; \boldsymbol{\mathbf{\theta_{joint}}_i}).$ To ensure the reliability of predicting the rotation matrix from estimating the plurality, we wish to minimize the log-likelihood distribution by defining the NLL loss with respect to the distribution of the Fisher matrix on the 3D joint rotation as follows: 
\begin{align} 
L_{joint} & = -\sum\nolimits_{n=1}^K \log \mathcal{M}\left(\mathbf{\theta_{joint}}_i ; \mathbf{R}_i\right)\\  &=-\sum\nolimits_{n=1}^K \log c\left(\mathbf{\theta_{joint}}_i\right)-\operatorname{tr}\left(\mathbf{\theta_{joint}}_i^{T} \mathbf{R}_i\right).
\end{align}

\section{Analysis of the Metrics}
\label{Sec. C}
\subsection{PSNR.}
Traditionally usage is to measure lossy image quality and video compression. A higher PSNR generally indicates better quality.

\begin{equation}
   PSNR = 10 \cdot \log_{10}\left(\frac{MAX_I^2}{MSE}\right),
\end{equation}
where $MAX_I$ is the maximum possible pixel value of the image, and Mean Squared Error (MSE) is the average of the squares of the pixel differences. PSNR emphasizes the fidelity aspect of the image, quantifying how accurately the image content is preserved after operations like compression. PSNR is a traditional metric that measures pixel-wise accuracy but might not always correlate well with perceived image quality. It is sensitive to absolute errors but may not align perfectly with human visual perception.
\subsection{SSIM.}
It quantifies the similarity between two images in terms of luminance, contrast, and structure. The values of SSIM range from -1 to 1, where 1 indicates perfect similarity.

\begin{equation}
    SSIM(x,y)=\frac{(2\mu_x\mu_y+c_1)(2\sigma_{xy}+c_2)}{(\mu_x^2+\mu_y^2+c_1)(\sigma_x^2+\sigma_y^2+c_2)}.
\end{equation}

SSIM is designed to provide a more perceptually relevant measure of image quality by incorporating texture and structural information. It is particularly useful in contexts where preserving structural integrity is crucial.
\subsection{LPIPS.}
It measures perceptual similarity and attempts to mimic human visual perception more closely than traditional metrics. LPIPS does not have a simple mathematical formula like SSIM or PSNR. It is computed using deep neural networks (e.g. VGG). LPIPS is focused on capturing the perceptual differences between images, including textures, colors, and patterns that traditional metrics might overlook. LPIPS leverages deep learning to provide a perceptual similarity measure, making it highly relevant for evaluating the output of generative models and other applications where human perception is the ultimate judge of quality. It's particularly useful for evaluating the performance of generative models where perceptual quality is more important than pixel-wise accuracy.

\textbf{In a word}, while traditional metrics like SSIM and PSNR continue to be valuable for many image processing tasks, especially those where exact reconstruction is crucial, \textbf{LPIPS represents an important shift towards prioritizing human perception} in the evaluation of image quality. This makes LPIPS particularly important in the current era of Artificial Intelligence and machine learning, where the end goal is often to generate or enhance images in ways that are \textbf{meaningful and satisfying to human observers}. Show in Table 1, our methodology not only excels in LPIPS, underscoring our emphasis on perceptual quality but also achieves state-of-the-art results, showcasing the superiority of our approach in aligning with human perceptual standards.

\begin{table*}[htbp]
\vspace{-5mm}
\centering  
 \caption{$512 \times 512$ results of each sub-datasets on ZJU-MoCap~\cite{NeuralBody_ZJU-MoCap} dataset and Monocap~\cite{InstantNVR} dataset.}
\fontsize{8}{15}\selectfont  
 
\begin{tabular}{l c c c c c c}
\hline
\hline 
        Method & PSNR↑\ \ & SSIM↑\ \ & LPIPS*↓ & PSNR↑\ \ & SSIM↑\ \ & LPIPS*↓ \\ \hline
    \multicolumn{7}{c}{MonoCap} \\ \hline
    ~ &~ & Lan   & ~ &~ & Marc  & ~ \\ \hline
    InstantNVR~\cite{InstantNVR} & 32.78 & \textbf{0.987} & 17.13 & 33.84 & \textbf{0.989} & 16.92 \\ \hline
    InstantAvatar~\cite{InstantAvatar} & 32.43 & 0.978 & 20.90 & 33.88 & 0.979 & 24.40 \\ \hline
    Human101(100s)~\cite{Li2023Human101T1} & 32.63 & 0.982 & 14.21 & 34.84 & 0.983 & 19.21 \\ \hline
    Human101(5min)~\cite{Li2023Human101T1} & 32.56 & 0.982 & 13.20 & 35.02 & 0.983 & 17.25 \\ \hline
    MOSS  & \textbf{34.43} & 0.986 & \textbf{9.66} & \textbf{35.82} & 0.986 & \textbf{12.69} \\ \hline
    ~ & ~ & Olek   & ~ &~ & Vlad  & ~ \\ \hline
    InstantNVR~\cite{InstantNVR} & 34.95 & \textbf{0.991} & 13.93 & 28.88 & \textbf{0.984} & 18.72 \\ \hline
    InstantAvatar~\cite{InstantAvatar} & 34.21 & 0.98 & 20.60 & 28.20 & 0.972 & 34.00 \\ \hline
    Human101(100s)~\cite{Li2023Human101T1} & 34.31 & 0.982 & 15.07 & \textbf{28.96} & 0.977 & 23.56 \\ \hline
    Human101(5min)~\cite{Li2023Human101T1} & 34.09 & 0.983 & 14.09 & 28.84 & 0.977 & 21.49 \\ \hline
    MOSS  & \textbf{35.35} & 0.987 & \textbf{9.01} & 28.65 & 0.980 & \textbf{15.35} \\ \hline \hline
    \multicolumn{7}{c}{ZJU-MoCap}\\ \hline
    ~ & ~ &377 &  ~ & ~ &386  & ~ \\ \hline
    InstantNVR~\cite{InstantNVR} & 31.69 & \textbf{0.981} & 32.04 & 33.16 & \textbf{0.979} & 38.67 \\ 
    InstantAvatar~\cite{InstantAvatar} & 29.90 & 0.961 & 49.00 & 30.67 & 0.917 & 111.5 \\ 
    Human101(100s)~\cite{Li2023Human101T1} & \textbf{32.18} & 0.977 & 24.65 & \textbf{33.94} & 0.972 & 36.03 \\ 
    Human101(5min)~\cite{Li2023Human101T1} & 32.02 & 0.976 & 21.35 & 33.78 & 0.969 & 33.73 \\ 
    MOSS  & 31.90 & 0.974 & \textbf{15.41} & 33.42 & 0.969 & \textbf{23.96} \\ \hline
    ~ & ~ &387  & ~ &~ & 392  & ~ \\ \hline
    InstantNVR~\cite{InstantNVR} & 27.73 & \textbf{0.961} & 55.90 & 31.81 & \textbf{0.973} & 39.25 \\ \hline
    InstantAvatar~\cite{InstantAvatar} & 27.49 & 0.928 & 86.30 & 29.39 & 0.934 & 96.90 \\ \hline
    Human101(100s)~\cite{Li2023Human101T1} & \textbf{28.32} & 0.956 & 47.76 & \textbf{32.22} & 0.966 & 41.89 \\ \hline
    Human101(5min)~\cite{Li2023Human101T1} & 28.26 & 0.956 & 44.57 & 32.11 & 0.967 & 39.23 \\ \hline
    MOSS  & 27.98 & 0.955 & \textbf{31.29} & 31.84 & 0.965 & \textbf{25.90} \\ \hline
    ~ &~ & 393   & ~ &~ & 394 & ~ \\ \hline
    InstantNVR~\cite{InstantNVR} & 29.46 & \textbf{0.964} & 46.68 & 31.26 & \textbf{0.969} & 39.89 \\ \hline
    InstantAvatar~\cite{InstantAvatar} & 28.17 & 0.931 & 86.60 & 29.64 & 0.943 & 64.20 \\ \hline
    Human101(100s)~\cite{Li2023Human101T1} & 29.69 & 0.957 & 46.52 & \textbf{31.37} & 0.967 & 40.16 \\ \hline
    Human101(5min)~\cite{Li2023Human101T1} & 29.52 & 0.956 & 44.15 & 31.25 & 0.968 & 36.86 \\ \hline
    MOSS  & \textbf{29.90} & 0.957 & \textbf{29.70} & 31.03 & 0.959 & \textbf{26.12} \\ \hline

\bottomrule
\end{tabular}
\label{tab:tab1}
\end{table*}

\section{More Experiments}
\label{Sec. D}
In this section, we present the exploratory experiments that we have conducted. These primarily include the rotation experiments of spherical harmonics coefficients and the data augmentation experiments aimed at addressing the issue of occlusions from a single view. Our exploration revealed that these techniques do not significantly enhance the reconstruction of clothed human bodies. Furthermore, we also show the qualitative and quantitative results that are not included in the main text. Table \ref{tab:tab1} displays the results of MOSS in comparison to other methods on each sub-dataset of ZJU-MoCap and MonoCap. MOSS demonstrates lower LPIPS* across all sub-datasets compared to other methods, indicating MOSS's superior performance in terms of human visual perception and stability. Figure \ref{fig:appendix_b} presents the qualitative results. Detailed analysis is as follows.

\begin{table}[htbp]
    \centering  
     \caption{Results of SH experiments for each subject on the ZJU-MoCap dataset.}
    \fontsize{8}{15}\selectfont 
    \resizebox{1\linewidth}{!}{
    \begin{tabular}{lcccccc}
    \hline
    \hline
        Method & PSNR↑\ \ & SSIM↑\ \ & LPIPS*↓ & PSNR↑\ \ & SSIM↑\ \ & LPIPS*↓ \\ \hline
        ~ & ~ & 377 & ~ & ~ & 386 & ~ \\ \hline
        MOSS  & \textbf{31.75} & 0.974 & 17.57 & \textbf{33.55} & \textbf{0.969} & \textbf{27.28} \\ \hline
        MOSS+SH \ \  \ \  \ \ \ & 31.65 & 0.974 & 17.57 & 32.80 & 0.966 & 29.62 \\ \hline
        ~ & ~ & 387 & ~ & ~ & 392 & ~ \\ \hline
        MOSS  & \textbf{28.11} & \textbf{0.956} & 36.22 & \textbf{31.56} & 0.965 & \textbf{28.43} \\ \hline
        MOSS+SH & 27.94 & 0.954 & \textbf{34.80} & 31.55 & 0.965 & 28.52 \\ \hline
        ~ & ~ & 393 & ~ & ~ & 394 & ~ \\ \hline
        MOSS  & \textbf{30.04} & \textbf{0.957} & 33.83 & \textbf{31.27} & \textbf{0.960} & 29.45 \\ \hline
        MOSS+SH & 29.72 & 0.956 & \textbf{31.88} & 30.94 & 0.959 & \textbf{28.59} \\ \hline
\bottomrule
\end{tabular}    
    }
\label{tab:tab2}
\end{table}

\subsection{Exploration for Rotating Spherical Harmonic Coefficients. }
In Section~\ref{sec:3.2} and Section~\ref{sec:3.3} of the main text, we have guided the 3D Gaussians with global motion information from the SMPL kinematic tree, which constrains the new Gaussians precisely and to the finest granularity. We use the joint rotation $R_{mode}$ as guidance to  Gaussian's rotation for the first time. During our experiments, we actually also attempt to explore the impact of motion guidance on spherical harmonic coefficients ~\cite{Hu2023GauHumanAG, Kerbl20233DGS}, with the aim of color refinement in a single view.
 The directional appearance component (color) of the radiance field is represented via Spherical Harmonics (SH). ~\cite{Hu2023GauHumanAG} has explored that directly rotate the spherical harmonic coefficients with a Wigner-D matrix. They find that rotating SH coefficients have little effect on the final performance. Since SH possesses \textbf{rotational invariance}, the independent variable can be rotated first before being passed into the SH function.

We use \textbf{the joint rotation $R_{mode}$ to guide the rotation of the spherical harmonic coefficients}. Due to the rotational invariance of SH,  a more effective approach is to inversely rotate the \textbf{view direction} $d_i$, which is computed by the final Gaussian position $x_{new}$ and the camera center. The rotated view direction:
\begin{equation}
    d'_i=R_{mode} \times d.
\end{equation}

The experimental results are shown in Table \ref{tab:tab2}. (Note that the hyperparameter setups are different from the final result in the main text.)

\subsection{Data Augmentation for Occlusion of Folds.}
\begin{table}[htbp]
    \centering  
     \caption{Data Augmentation experiments for each subject on the ZJU-MoCap dataset.}·
    \fontsize{8}{15}\selectfont  
    \resizebox{1\linewidth}{!}{
     \begin{tabular}{lcccccc}
    \hline
    \hline
        Method & PSNR↑\ \ & SSIM↑\ \ & LPIPS*↓ & PSNR↑\ \ & SSIM↑\ \ & LPIPS*↓ \\ \hline
        ~ & ~ &377 &  ~ & ~ &386  & ~ \\ \hline
        MOSS  & 31.90 & 0.974 & \textbf{15.41} & 33.42 & 0.969 & \textbf{23.96} \\ \hline
        MOSS+DataAug\ \  & \textbf{31.91} & 0.974 & 15.75 & \textbf{33.46} & 0.969 & 24.48 \\ \hline
        ~ &~ & 387  & ~ & ~ &392   & ~ \\ \hline
        MOSS  & 27.98 & 0.955 & \textbf{31.29} & 31.84 & 0.965 & \textbf{25.90} \\ \hline
        MOSS+DataAug & \textbf{28.07} & 0.955 & 32.11 & \textbf{31.86} & 0.965 & 26.57 \\ \hline
        ~ &~ & 393   & ~ &~ & 394   & ~ \\ \hline
        MOSS  & 29.90 & 0.957 & \textbf{29.70} & \textbf{31.03} & 0.959 & \textbf{26.12} \\ \hline
        MOSS+DataAug & \textbf{29.94} & 0.957 & 30.23 & 31.02 & 0.959 & 26.82 \\ \hline
\bottomrule
\end{tabular}   
    }
\label{tab:tab3}
\end{table}
In the course of our experiments, we found that the input from a single, fixed camera leads to issues with occlusion of folds from different viewpoints, resulting in weak supervision. To mitigate this issue, we developed a data augmentation technique that enhances the diversity of camera perspectives. Existing approaches often rotate the camera pose while keeping the SMPL global orientation fixed. In contrast, we now shift to fixing the camera pose and rotating the SMPL global orientation instead. Through data augmentation, we simulate the effect of the camera rotating around the human subject, thereby generating multiple camera poses. The results are shown in Table \ref{tab:tab3}.

Although there is an improvement in PSNR, it comes at the cost of worsening LPIPS scores. Upon careful consideration, we do not implement the data augmentation technique. We adopt the UID strategy outlined in Section 3.4 of the main text as our solution to address this issue effectively.

\subsection{More Qualitative Results.}

\begin{figure}[h]
    \centering
    \includegraphics[width=0.83\linewidth]{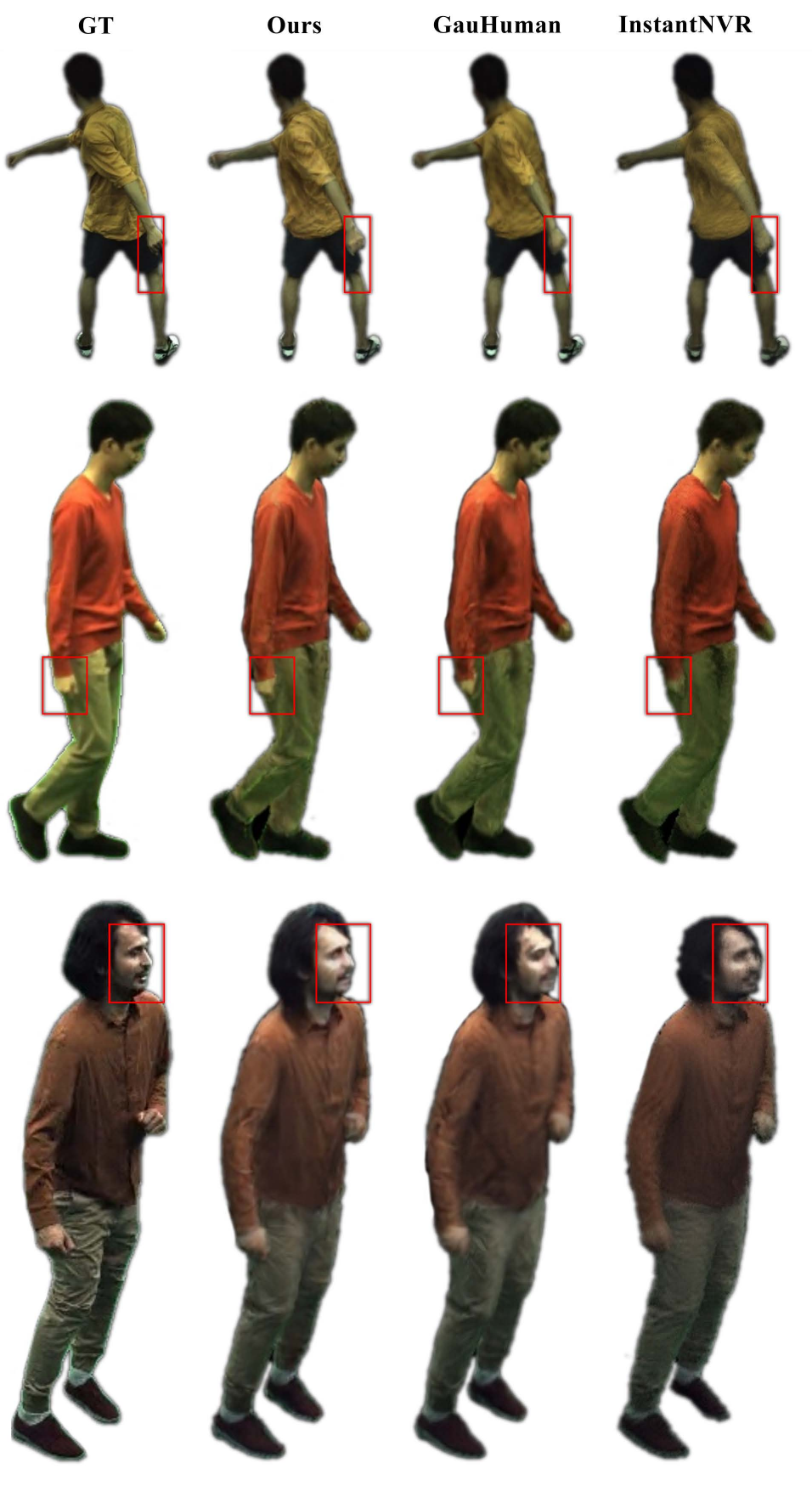}
    \caption{Rendered images with boxes highlighting the details of the reconstructed and ground truth human bodies.}
\label{fig:appendix_b}
\end{figure}

More comparative experimental results is shown in Figure \ref{fig:appendix_b}, featuring rendered images from various sub-datasets of ZJU-Mocap and MonoCap. The figure highlights the detailed reconstruction of the human body by MOSS, showcasing superior detail in human joints and clearly demarcated edges of clothing and joints (e.g., the shorts and legs in the first row). Furthermore, our method excels in rendering realistic details of fists and faces.

\end{CJK}
\end{document}